\documentclass{article}
\usepackage{spconf,amsmath,graphicx,hyperref}

\usepackage{amssymb}
\usepackage{booktabs}
\usepackage{array}
\usepackage{pifont}
\usepackage{multirow}
\usepackage{colortbl}
\usepackage{xcolor}
\usepackage{float}

\title{CoVA: Text-Guided Composed Video Retrieval for Audio-Visual Content}
\name{Gyuwon Han$^{1*}$, Young Kyun Jang$^{2*}$, Chanho Eom$^{1\dagger}$%
\thanks{$^{*}$ Equal contribution. $^{\dagger}$ Corresponding author.}}
\address{$^{1}$Chung-Ang University \{jshfu, cheom\}@cau.ac.kr,\\ $^{2}$Google DeepMind \{kyun0914@gmail.com\}}

\begin{document}
%
\maketitle
\begin{abstract}
Composed Video Retrieval (CoVR) aims to retrieve a target video from a large gallery using a reference video and a textual query specifying visual modifications. However, existing benchmarks consider only visual changes, ignoring videos that differ in audio despite visual similarity. To address this limitation, we introduce \textbf{Co}mposed retrieval for \textbf{V}ideo with its \textbf{A}udio (\textbf{CoVA}), a new retrieval task that accounts for both visual and auditory variations. To support this, we construct AV-Comp, a benchmark consisting of video pairs with cross-modal changes and corresponding textual queries that describe the differences. We also propose AVT Compositional Fusion (AVT), which integrates video, audio, and text features by selectively aligning the query to the most relevant modality. AVT outperforms traditional unimodal fusion and serves as a strong baseline for CoVA. Examples from the proposed dataset, including both visual and auditory information, are available at \url{https://perceptualai-lab.github.io/CoVA/}.

\end{abstract}

\section{Introduction}
\label{sec:intro}
\begin{figure}[!t]
    \centering
    \includegraphics[width=0.48\textwidth]{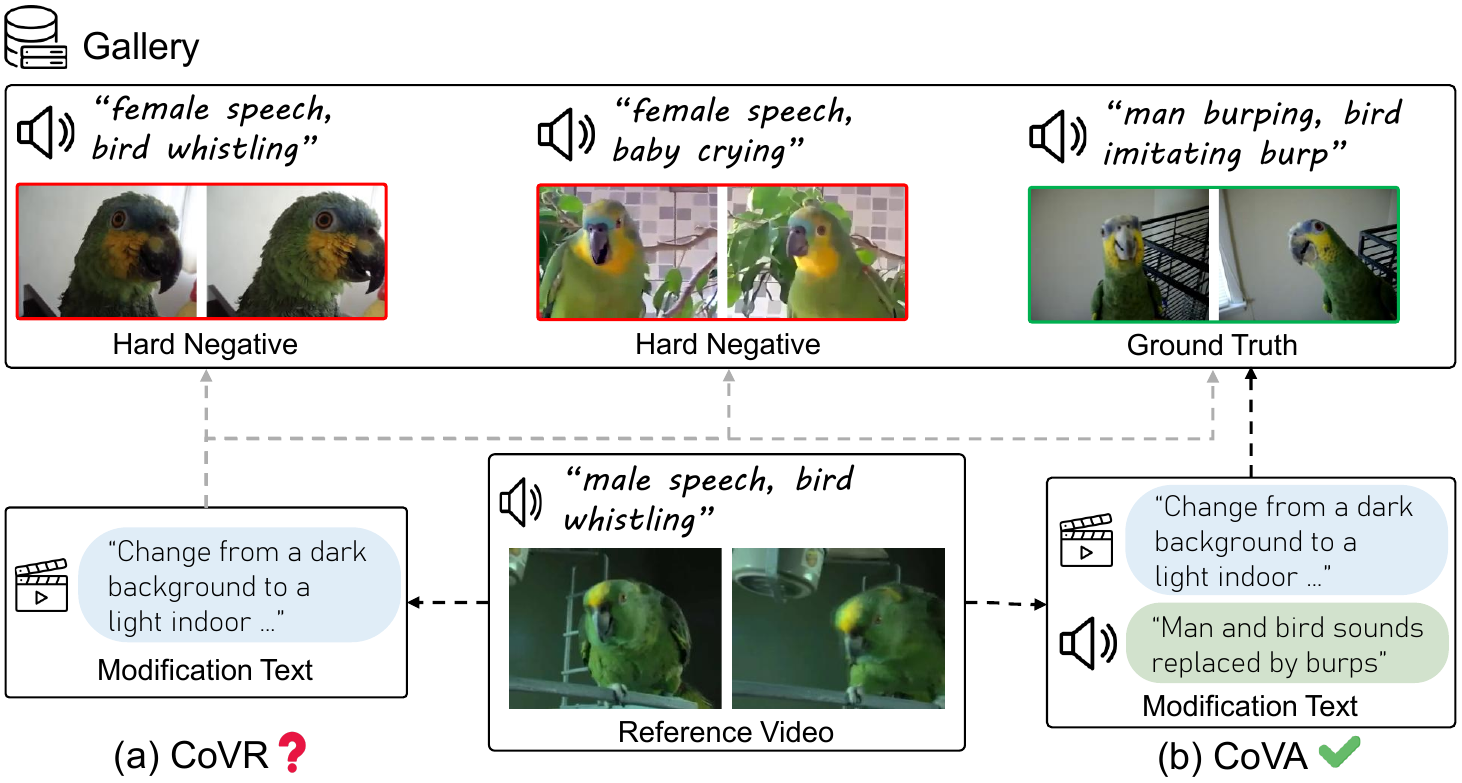}
    \vspace{-1.5em}
    \caption{While the (a) existing CoVR considers only visual modifications, (b) CoVA utilizes both visual and auditory information to support more fine-grained retrieval.}
    \label{fig:fig1} 
    \vspace{-1em}
\end{figure}

Recent advances in video-text retrieval have increasingly recognized the importance of audio, leading to the development of various methods for effectively fusing audio-visual signals~\cite{lin2022eclipse, ibrahimi2023audio, jeong2025learning}.
A major paradigm shift in this domain is the emergence of large-scale multimodal models that aim to unify textual, auditory, and visual data into a single embedding space~\cite{girdhar2023imagebind, chen2023vast, wang2024internvideo2, cicchetti2025gramian,liu2024valor}. Meanwhile, composed video retrieval (COVR) (Fig.~\ref{fig:fig1}(a)) has gained attention as a novel retrieval paradigm that goes beyond conventional semantic matching, enabling users to modify a reference video using a textual query to retrieve a desired target.
Although this task is initially explored in the image domain~\cite{saito2023pic2word, baldrati2023zero}, recent methods have extended it to video, leveraging large multimodal models with semi-supervised training ~\cite{jang2024visual} and adaptive aggregation networks for cross-modal fusion ~\cite{jang2024spherical,huang2025median}.
Such approaches are further scaling to video-level datasets via automated caption mining, as demonstrated in CoVR~\cite{ventura2024covr}, as well as through discriminative embedding learning~\cite{thawakar2024composed} and benchmark design for fine-grained egocentric tasks~\cite{hummel2024egocvr}. However, existing studies remain largely constrained to the visual modality, often neglecting the impact of audio changes on user intent.
For instance, current models assume that two visually similar videos are semantically equivalent, even though they may contain different audio. In practical scenarios, users may perceive these videos as semantically different due to their audio, which poses challenges for visually-grounded models to capture user intent in multimodal settings. To address this limitation, we introduce Composed retrieval for Video with its Audio (CoVA), a new task that incorporates both visual and auditory changes through textual queries (Fig.~\ref{fig:fig1}(b)). The main contributions of this work are:
\vspace{-0.2em}
\begin{itemize}
    \item We introduce AV-Comp, the first benchmark for composed video retrieval with aligned audio-video changes.
    \item We propose CoVA with Gated Fusion Transformer and AVT Compositional Fusion, consistently outperforming arithmetic fusion.
    \item Through experiments on the newly proposed AV-Comp benchmark, we show that CoVA achieves state-of-the-art performance and outperforms baselines.
\end{itemize}
\vspace{-1.0em}

\begin{figure}[t!]
    \centering
\includegraphics[width=1\linewidth]{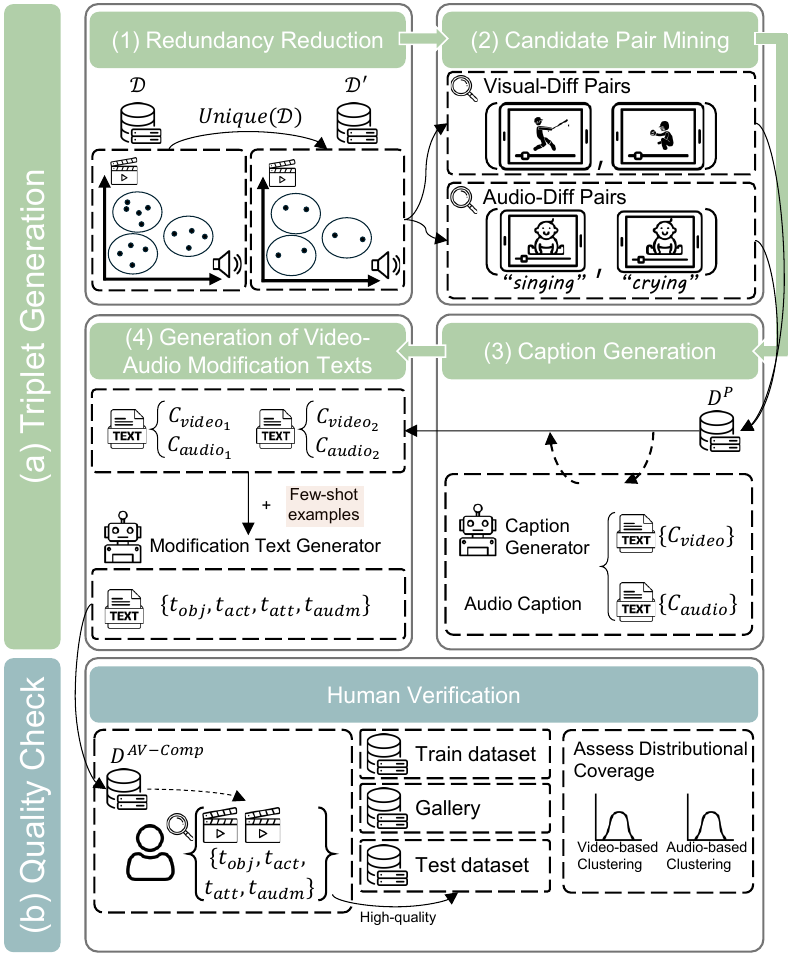}
\vspace{-1em}
    \caption{Overview of the dataset construction pipeline.}
    \label{fig:my_figure2}
\end{figure}

\section{Dataset construction}
\label{sec:Dataset construction}
We propose a triplet formulation that incorporates both visual and auditory information for multimodal retrieval. Each triplet takes the form \(((v^q, a^q), t, (v^t, a^t))\), where \(v\) denotes a sequence of image frames, \(a\) denotes the corresponding audio signal, and \(t\) denotes the textual modification query. The superscript \(q\) refers to the query, and \(t\) refers to the target. To ensure the quality of audio captions, we use human-annotated captions from the AudioCaps 2.0 dataset~\cite{kim2019audiocaps}.
\vspace{-1.0em}

\paragraph*{Redundancy reduction.}
To avoid redundancy when both video and audio contents are highly similar, we treat such pairs $(v,a)$ as duplicates and retain only one instance (Fig.~\ref{fig:my_figure2}(\\a.1)). We denote original dataset as $\mathcal{D}$ and the deduplicated set as $\mathcal{D}' = \text{Unique}(\mathcal{D})$. To this end, we obtain the video embedding $E_v$ by averaging CLIP~\cite{radford2021learning} embeddings over 8 uniformly sampled frames from each video, and we extract the audio-caption embedding $E_a$ using the CLIP text encoder. Unique removes duplicates by retaining only one pair when $S(E_v) > 0.92$ and $S(E_a) > 0.96$, where $S(\cdot)$ denotes the cosine similarity between embeddings.
\vspace{-1.0em}

\paragraph*{Candidate pair mining.}
We collect two types of video pairs (Fig.~\ref{fig:my_figure2}(a.2)): 
(1) pairs that are visually similar but differ in their audio captions 
$(0.92 < S(E_v) < 0.96 \;\cup\; 0 < S(E_a) < 0.85)$, 
and (2) pairs that exhibit slight visual differences but share similar audio captions 
$(0.85 < S(E_v) < 0.88 \;\cup\; 0.95 < S(E_a) < 1)$ . We store these pairs in \( D^p \). 
We determine the thresholds for visual and audio-caption similarity by manually inspecting 100 videos and 100 audio captions, respectively.
\vspace{-1.0em}

\paragraph*{Video Caption Generation.}
To generate modification texts, both video captions and audio captions are required. We use Qwen2.5-VL-32B-Instruct~\cite{Qwen2.5-VL} to obtain video captions \( C_{\text{video}} \), and each caption contains three fields: object, action, and attribute (Fig.~\ref{fig:my_figure2}(a.3)). For audio captions \( C_{\text{audio}} \), we use human-annotated annotations from AudioCaps 2.0~\cite{kim2019audiocaps}. We generate these captions for each video pair selected from the candidate pool \( D^p \).
\vspace{-1.0em}

\paragraph*{Video-Audio modification text generation.}
To generate textual descriptions of video and audio differences between query-target pairs, we use Gemini~\cite{comanici2025gemini} with video captions \( C_{\text{video}_1}, C_{\text{video}_2} \), human-annotated audio captions \( C_{\text{audio}_1}, C_{\text{audio}_2} \), and several manually curated few-shot examples of video-audio differences (Fig.~\ref{fig:my_figure2}(a.4)). These pairs differ in either the visual or audio modality.  
Based on these inputs, the model generates structured difference descriptions across four aspects: object (\( t_{\text{obj}} \)), action (\( t_{\text{act}} \)), attribute (\( t_{\text{att}} \)), and audio (\( t_{\text{audm}} \)). We store the resulting video pairs and modification texts in \(D^{\text{AV-Comp}}\).
\vspace{-1.0em}

\paragraph*{Human verification.}
To construct the test set, we conduct human verification on each query-target video pair and its corresponding modification text in \( D^{\text{AV-Comp}} \) (Fig.~\ref{fig:my_figure2}(b)).
 We retain a sample only if it (1) shows a perceptually meaningful difference, (2) has a modification text that accurately and comprehensively captures the change across the four aspects (object, action, attribute, and audio), and (3) contains no hallucinations in the video/audio captions or the difference description. We include high-quality samples that pass this verification in the final test set. We assign the remaining samples to the training set or use them to construct a gallery set of 1,000 additional videos. We use the gallery separately, which enlarges the candidate pool and raises retrieval difficulty. In total, we build a dataset with 8,357 training triplets, 1,001 test triplets, and 1,000 additional gallery videos. 

\section{Method}
\begin{figure}[t!]
    \centering
    \includegraphics[width=0.9\linewidth]{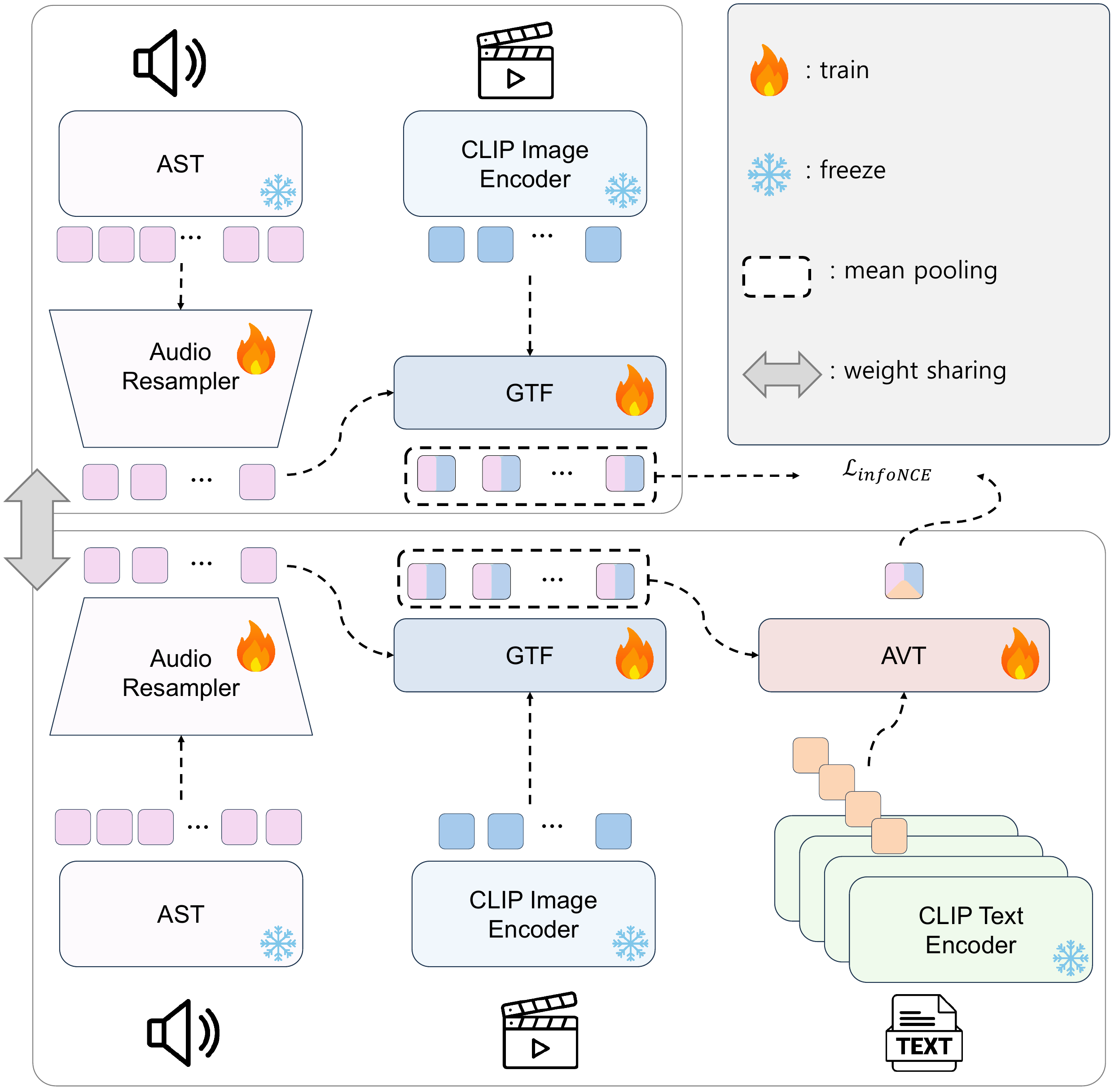}
    \caption{Overview of the proposed CoVA framework.}
    \label{fig:my_figure3}
\end{figure}

Our overall framework is illustrated in Fig.~\ref{fig:my_figure3} and consists of three modules: Feature Extraction, Gated Fusion Transformer (GFT), and AVT Compositional Fusion (AVT). Feature Extraction extracts features from video, text, and audio inputs. GFT fuses video and audio features into unified multimodal representations, and AVT integrates the modification text with the fused audio–video representation to generate the final query representation.
\vspace{-1.0em}

\label{sec:method}
\paragraph*{Feature Extraction.}
From an input video $v$, we uniformly sample frames and extract the [CLS] output of CLIP~\cite{radford2021learning} image encoder to form the frame embedding set $f=[f_1,\ldots,f_N]$, where $N$ denotes the number of sampled frames. 
The modification query $t$ is structured into four aspects, and we independently encode each field using CLIP text encoder, taking the [EOS] output as the representation to obtain $t_{obj}, t_{act}, t_{att}, t_{audm} \in \mathbb{R}^D$. 
We convert the audio signal $a$, aligned with the video, into a log-Mel spectrogram, process it with the AST~\cite{gong2021ast} encoder, and pass it through a query-based resampler to reduce the token length. This produces the final embeddings $a=[a_1,\ldots,a_M]$, where $M$ denotes the number of learnable tokens.
\vspace{-1.0em}

\paragraph*{Gated Fusion Transformer.}
Motivated by~\cite{jeong2025learning}, we adopt the Gated Fusion Transformer (GFT) to integrate visual and auditory features into a unified representation. The GFT processes $f \in R^{N \times D}$ and $a \in R^{M \times D}$ through $L$ cross-attention layers and generates refined visual embeddings $f^{(L)} \in R^{N \times D}$ that incorporate complementary audio cues. We then apply mean pooling over the final layer outputs to obtain a fixed-length fused embedding, defined as $f_{av} = \text{MeanPool}(f^{(L)}) \in R^D$.
\vspace{-1.0em}

\paragraph*{AVT Compositional Fusion.}
To generate the final representation $f_{avt}$ by combining the audio-video feature $f_{av}$ with the modification text, we introduce a simple yet effective module, AVT Compositional Fusion (AVT). AVT uses the audio-video fused embedding $f_{av}$ and four independently encoded text components $(t_{obj}, t_{act}, t_{att}, t_{audm})$ as inputs. It dynamically adjusts the contribution of each component according to the semantic intent of the query, unlike averaging approaches. We first concatenate the multimodal query embeddings into a single vector and pass it through a simple MLP to predict five weights. Each weight $w_i \in [0,1]$ is normalized via a sigmoid function. We apply the predicted weights to each component and compute the final representation as:
\begin{equation}
f_{avt} = \sum w_i f_i, \quad f_i \in \{ f_{av}, t_{obj}, t_{act}, t_{att}, t_{audm} \}.
\label{eq:agt}
\end{equation} We apply L2 normalization to $f_{avt}$ before using it in retrieval. We train this design to adaptively emphasize the most relevant modalities or semantic components based on the input query.

\paragraph*{Training Objective.}  
We train the model with a symmetric InfoNCE loss~\cite{oord2018representation}. 
We represent the query by the fused query embedding $f_{avt}$, denoted as $q_i$, 
and the target by the fused target embedding $f_{av}$, denoted as $t_i$. 
We compute the final loss as
\begin{equation}
\label{eq:infonce}
\begin{aligned}
\mathcal{L}_{\text{InfoNCE}} = - \frac{1}{2N} \sum_{i=1}^N \Bigg[
& \log \frac{\exp\big((q_i^\top t_i)/\tau\big)}{\sum_{j=1}^N \exp\big((q_i^\top t_j)/\tau\big)} \\
& {}+ \log \frac{\exp\big((t_i^\top q_i)/\tau\big)}{\sum_{j=1}^N \exp\big((t_i^\top q_j)/\tau\big)}
\Bigg],
\end{aligned}
\end{equation}
where $\tau$ denotes a learnable temperature parameter.

\section{EXPERIMENTS}
In this section, we present the experimental results of the proposed CoVA framework. 
We begin by describing the experimental setup, including the dataset, evaluation metrics, and implementation details, in Section~\ref{sec:experimental setup}. We then report the main results on our audio and video benchmark in Section~\ref{sec:results on AV-comp} and analyze the contribution of each component through ablation studies in Section~\ref{sec:ablation}.   

\subsection{Experimental Setup}
\label{sec:experimental setup}
\paragraph*{Dataset.} We train our model on the training set of the AV-Comp benchmark and evaluate it on the test set that contains human-verified video-audio pairs with modification instructions spanning both visual and auditory variations.
\vspace{-1.0em}

\paragraph*{Evaluation Metrics.}
We evaluate using R@1/5/10 and MnR. R@K is the percentage of queries where the correct target appears in the top K results, and MnR is the average rank of the correct target.
\vspace{-1.0em}

\paragraph*{Implementation Details.}
We use CLIP (ViT-B/32) as the visual and text encoder, and AST~\cite{gong2021ast}, pre-trained on ImageNet~\cite{deng2009imagenet} and AudioSet~\cite{gemmeke2017audio}, as the audio encoder. All encoders are kept frozen, and we train only the Gated Fusion Transformer (GFT) and the AVT Compositional Fusion (AVT) module. We train the model for 10 epochs with a batch size of 64 and a learning rate of $1 \times 10^{-4}$ on four NVIDIA RTX 4090 GPUs.

\begin{table}[!t]
\caption{Retrieval performance of models fine-tuned on the AV-Comp benchmark with various modality combinations and fusion strategies.}
\centering
\resizebox{0.95\linewidth}{!}{%
\begin{tabular}{lccccc}
\toprule
\textbf{Input} & \textbf{Fusion} & \textbf{R@1\ensuremath{\uparrow}} & \textbf{R@5\ensuremath{\uparrow}} & \textbf{R@10\ensuremath{\uparrow}} & \textbf{MnR\ensuremath{\downarrow}} \\
\midrule
Random     & -     & 0.06  & 0.32  & 0.62  & 935.8 \\
\midrule
T          & -     & 19.7  & 44.9  & 60.5  & 19.9  \\
V          & -     & 21.5  & 49.7  & 65.3  & 21.4  \\
A          & -     & 1.0   & 1.8   & 3.9   & 542.8 \\
\midrule
V,A        & GFT     & 22.3  & 52.3  & 68.9  & 16.2  \\
V,T        & Avg     & 28.8  & 64.3  & 78.8  & 10.8  \\
A,T        & Avg     & 22.2  & 53.5  & 69.4  & 13.1  \\
\midrule
T,V,A      & Avg + Avg   & {25.9}  & {60.7}  & {75.2}  & {11.1}  \\
T,V,A      & Avg + AVT   & 28.1  & 63  & 77  & \underline{9.4}  \\
T,V,A      & GFT + Avg   & \underline{30.4}  & \underline{65.7}  & \underline{80.0}  & 10.5  \\
T,V,A      & GFT + AVT   & \textbf{31.4} & \textbf{66.0} & \textbf{80.5} & \textbf{9.3} \\
\bottomrule
\end{tabular}%
}
\label{tab:CoVA_test}
\end{table}

\subsection {Results on AV-Comp Benchmark}
\label{sec:results on AV-comp}
We conduct a comprehensive evaluation of baselines on the AV-Comp benchmark, and summarize the results in Table.~\ref{tab:CoVA_test}. The first block reports the performance of a random baseline that selects targets at random, independent of the query. The second block presents unimodal retrieval results using only one modality among text (T), video (V), or audio (A). The third block evaluates query settings that combine two modalities. For the video and audio combination, we apply the Gated Fusion Transformer (GFT) to integrate the two modalities, as they originate from the same video clip and require deeper cross-modal interaction. In contrast, for the video+text and audio+text combinations, we simply average the outputs from each encoder due to their modality-level heterogeneity. The fourth block considers query settings that involve all three modalities (video, audio, and text). In this setting, we apply two-stage fusion where the first term denotes the audio-video fusion method and the second term denotes the method for integrating text with the fused audio-video representation. We compare GFT and AVT against simple averaging (Avg) for audio-video fusion and text integration, respectively. From these experimental results, we draw the following conclusions.
(1) While audio alone performs poorly as a query modality, it significantly improves retrieval performance when combined with text or video, demonstrating its effectiveness as a complementary modality.
(2) With naive averaging, adding audio leads to marginal improvements, showing little difference between the T+V and T+V+A configurations.
(3) The AVT module, which employs query-aware weighted fusion, slightly outperforms naive averaging, yielding a 1.2-point improvement in MnR. This result demonstrates that AVT, despite its simple structure, effectively integrates information across multiple modalities. In Table~\ref{tab:model_comparison_ft}, we compare our method with LanguageBind~\cite{zhu2023languagebind} and ImageBind~\cite{girdhar2023imagebind} under the same fine-tuning setting. In this experiment, we use CLIP-L to match LanguageBind’s model size, and use AST as the audio encoder. Since both baselines lack a module for multi-modal composition, we apply the same GFT+AVT to all methods and train only GFT+AVT while freezing the encoders. As a result, our method achieves superior performance on AV-Comp, attaining 35.9\% R@1 and surpassing other multimodal foundation models.

\begin{table}[!t]
\caption{Performance on our dataset compared with baseline and existing multimodal models.}
\centering
\resizebox{0.85\linewidth}{!}{%
\begin{tabular}{lcccc}
\toprule
\textbf{Method} & \textbf{R@1\ensuremath{\uparrow}} & \textbf{R@5\ensuremath{\uparrow}} & \textbf{R@10\ensuremath{\uparrow}} & \textbf{MnR\ensuremath{\downarrow}} \\
\midrule
ImageBind~\cite{girdhar2023imagebind}        & 20.2  & 50.5  & 65.4  & 14.6 \\
LanguageBind~\cite{zhu2023languagebind}     & \underline{27.17} & \underline{61.44} & \underline{77.12} & \underline{8.7}  \\
CoVA (Ours)                                       & \textbf{35.9} & \textbf{73.7} & \textbf{86.4} & \textbf{6.2} \\

\bottomrule
\end{tabular}%
}
\label{tab:model_comparison_ft}
\end{table}

\subsection {Ablation studies}
\label{sec:ablation}
Table.~\ref{tab:ablation} presents the ablation study to verify the contribution of the four textual components $(t_{\text{obj}}, t_{\text{act}}, t_{\text{att}}, t_{\text{audm}})$. Removing any single component consistently leads to performance degradation across all metrics. This demonstrates that each of four components contributes significantly to overall retrieval performance. The results confirm that the proposed textual query structure leverages complementary information from each component to enhance multimodal alignment.
\begin{table}[H]
\caption{Ablation study on the contribution of each textual query component.}
\centering
\resizebox{0.75\linewidth}{!}{%
\begin{tabular}{lcccc}
\toprule
\textbf{Setting} & \textbf{R@1\ensuremath{\uparrow}} & \textbf{R@5\ensuremath{\uparrow}} & \textbf{R@10\ensuremath{\uparrow}} & \textbf{MnR\ensuremath{\downarrow}} \\
\midrule
w/o $t_{\text{obj}}$   & 26.8 & 62.2 & 75.8 & 9.7 \\
w/o $t_{\text{act}}$   & \underline{30.9} & 64.6 & 78.7 & 10.8 \\
w/o $t_{\text{att}}$   & 28.8 & 63.3 & 77.5 & 10.9 \\
w/o $t_{\text{audm}}$  & 30.7 & \textbf{66.7} & \underline{80.3} & \underline{9.6} \\
\midrule
CoVA (Ours)         & \textbf{31.4} & \underline{66.0} & \textbf{80.5} & \textbf{9.3} \\
\bottomrule
\end{tabular}%
}
\label{tab:ablation}
\end{table}

\section{CONCLUSION}
\label{sec:conclusion}
We have proposed AV-Comp, the first benchmark for composed video retrieval that explicitly accounts for both visual and auditory differences. In addition, we have introduced AVT Compositional Fusion (AVT), a method designed to effectively handle complex cross-modal variations. It has demonstrated improved performance over existing baselines, and further gains are expected when it is replaced with a more sophisticated fusion module. Experimental results have shown that AV-Comp not only provides diverse content and high-quality annotations, but also serves as a valuable diagnostic tool for evaluating how well pretrained multimodal models align audio, video, and text representations.

\section*{Acknowledgements}
This work was partly supported by the Technology development Program(RS-2025-25442747) funded by the Ministry of SMEs and Startups(MSS, Korea), and the MSIT(Ministry of Science and ICT), Korea, under the Graduate School of Metaverse Convergence support program(IITP-2024-RS-2024-00418847) supervised by the IITP(Institute for Information \& Communications Technology Planning \& Evaluation.



\small\bibliographystyle{IEEEbib}
\clearpage
\bibliography{strings,refs}

\end{document}